\documentclass[nohyperref]{article}

\usepackage{microtype}
\usepackage{graphicx}
\usepackage{subfigure}
\usepackage{booktabs} % for professional tables

\usepackage{hyperref}

\usepackage[accepted]{icml2022}

\usepackage{amsmath}
\usepackage{amssymb}
\usepackage{mathtools}
\usepackage{amsthm}

\usepackage[capitalize,noabbrev]{cleveref}

\theoremstyle{plain}

\theoremstyle{definition}

\theoremstyle{remark}

\usepackage[textsize=tiny]{todonotes}

\icmltitlerunning{Explanation of Machine Learning Models of Colon Cancer Using SHAP Considering Interaction Effects}

\begin{document}

\twocolumn[
\icmltitle{Explanation of Machine Learning Models of Colon Cancer \\ Using SHAP Considering Interaction Effects}



\icmlsetsymbol{equal}{*}

\begin{icmlauthorlist}
\icmlauthor{Yasunobu Nohara}{kuma}
\icmlauthor{Toyoshi Inoguchi}{supp}
\icmlauthor{Chinatsu Nojiri}{ku}
\icmlauthor{Naoki Nakashima}{ku}
\end{icmlauthorlist}

\icmlaffiliation{kuma}{Faculty of Advanced Science and Technology, Kumamoto University, Kumamoto, Japan}
\icmlaffiliation{supp}{Fukuoka Health Promotion Support Center, Fukuoka, Japan}
\icmlaffiliation{ku}{Kyushu University Hospital, Fukuoka, Japan}

\icmlcorrespondingauthor{Yasunobu Nohara}{nohara@cs.kumamoto-u.ac.jp}

\icmlkeywords{Machine Learning, ICML}

\vskip 0.3in
]



\printAffiliationsAndNotice{}  % leave blank if no need to mention equal contribution

% Abstracts must be a single paragraph, ideally between 4-6 sentenceslong.
\begin{abstract}
When using machine learning techniques in decision-making processes, the interpretability of the models is important. Shapley additive explanation (SHAP) is one of the most promising interpretation methods for machine learning models. Interaction effects occur when the effect of one variable depends on the value of another variable. Even if each variable has little effect on the outcome, its combination can have an unexpectedly large impact on the outcome. Understanding interactions is important for understanding machine learning models; however, naive SHAP analysis cannot distinguish between the main effect and interaction effects. In this paper, we introduce the Shapley-Taylor index as an interpretation method for machine learning models using SHAP considering interaction effects. We apply the method to the cancer cohort data of Kyushu University Hospital (N=29,080) to analyze what combination of factors contributes to the risk of colon cancer.
\end{abstract}

\section{Introduction}
In recent years, remarkable breakthroughs have been achieved in machine learning technology, as typified by deep neural networks. Such technologies are expected to be used for decision-making in medical fields. In decision-making, it is essential to recognize why decisions are made. Although complex machine learning models such as deep learning and ensemble models can achieve high accuracy, they are more difficult to interpret than simple models such as linear models.
SHAP (SHapley Additive exPlanation)~\cite{lundberg2017unified} is a method for interpreting the results of machine learning by computing the contribution of each feature.
SHAP enables us to illustrate nonlinear relationships between features and the outcome.
SHAP is also highly compatible with linear models and the derivative of the SHAP value corresponds to the regression coefficient of the linear model.

An interaction effect occurs when the impact of one feature depends on another feature.
Even if each variable has little or no effect on the outcome, its combination can have an unexpectedly large impact on the outcome.
For example, potassium supplements and ACE inhibitors are safe alone; however, their combined use can cause hyperkalemia by drug interaction.

Understanding interactions is important for understanding machine learning models; however, naive SHAP analyses can only evaluate which features are important and cannot evaluate the effects of the main effect and interactions separately.

In this paper, we introduce the Shapley-Taylor index, proposed by Sundararajan et al., as a method for interpreting machine learning models. The index can separate the SHAP value into the effects of single features and the interactions.
The method is applied to the cancer cohort data of Kyushu University Hospital (N=29,080) to analyze what combination of factors contributes to the risk of colon cancer.

\section{Background}

\subsection{Shapley Additive Explanation}

\textit{SHapley Additive exPlanation} (SHAP)~\cite{lundberg2017unified} is a method for interpreting the results of machine learning by computing the contribution of each feature and represents the outcome of patient-$j$: $f(x^{(j)})$ as the sum of each features-$i$'s contribution $\phi_i(x_i^{(j)})$.
\begin{equation}
f(x^{(j)})=\phi_0+\sum_{i=1}^{K} \phi_i(x_i^{(j)})
\end{equation}

\begin{equation}
\phi_0=\frac{1}{N} \sum_{j=1}^{N} f(x^{(j)})
\end{equation}
\begin{equation}
\phi_i(x_i^{(j)})=\Phi(x_i^{(j)}) - \frac{1}{N} \sum_{k=1}^{N} \Phi({x_i}^{(k)})
\label{phi}
\end{equation}
, where $N$ is the number of patients and $\Phi(x_i)$ is the Shapley value for $x_i$.

The Shapley value is a fair profit allocation among many stakeholders depending on their contribution~\cite{roth1988shapley} and was derived from the name of the economist who introduced it. The Shapley value is defined as follows:
\begin{multline}
\Phi(x_i) \\ =\sum_{S \subseteq \{1,\cdots,K\} \setminus \{i\}} \frac{|S|!(K-|S|-1)!}{K!}[f_{x}(S\cup\{i\})-f_{x}(S)]
\label{SHAPdef}
\end{multline}
, where $K$ is the number of stakeholders or features.
The meaning of the bracket part of Eq.~(\ref{SHAPdef}) is that the contribution of entity-$i$ can be defined as a marginal contribution, i.e. the difference between the profit obtained by group-$S$ members only: $f_{x}(S)$ and that of both entity-$i$ and the group members: $f_{x}(S\cup\{i\})$.

The Shapley values can be calculated as long as the function value is defined even if the calculation method is a black box, 
The Shapley value is the only profit allocation method that satisfies the following four properties: efficiency, symmetry, linearity, and null player.

The computation time of naive SHAP calculations increases exponentially with the number of features $K$; however, Lundberg et al. proposed a polynomial-time algorithm for decision trees and ensemble trees model~\cite{lundberg2020local}.
This algorithm is integrated into major ensemble tree frameworks like XGBoost~\cite{chen2016xgboost} and LightGBM~\cite{ke2017lightgbm}.
Several studies have applied machine learning algorithms for medical data analysis and interpreted by SHAP~\cite{lundberg2020local,moncada2021explainable,inoguchi2021association}

Notara et al. proposed using the variance (L2-norm) of the SHAP value for measuring variable importance~\cite{nohara2022shap}.
The ranking result sorted by the absolute value of the beta coefficients $\beta_i$ in the generalized linear regression model is exactly the same as that of this definition.

In order to understand what the SHAP value means, we suppose the three-variable function $F(x,y,z)$ is expressed as the following equation.
\begin{equation}
F(x,y,z)=f_x(x)+f_y(y)+f_z(z)+g_{xy}(x,y)+g_{xz}(x,z)
\label{3value}
\end{equation}
, where $f(x)$ is the main effect term for the feature $x$ and $g(x,y)$ is the interaction term of features $x$ and $y$.

The respective contributions of each feature $x$, $y$, and $z$ in this function are as follows:
$$\phi(x)=f_x(x)+\frac{g_{xy}(x,y)+g_{xz}(x,z)}{2}$$
$$\phi(y)=f_y(y)+\frac{g_{xy}(x,y)}{2}$$
$$\phi(z)=f_z(z)+\frac{g_{xz}(x,z)}{2}$$
$$F(x,y,z)=\phi(x)+\phi(y)+\phi(z)$$
In other words, the SHAP value of the feature evaluates the sum of the main effect term of the feature and the interactions between the feature and others.
Therefore, we can evaluate which features are important; however, we cannot distinguish whether the main effect term is affecting the outcome or interactions are affecting the outcome.

\subsection{Shapley-Taylor index}
The Shapley-Taylor index is an extended version of the Shapley value~\cite{sundararajan2020shapley} and decomposes the SHAP value $\Phi(x_i) $ into the main term $\Phi(x_i,x_i)$ and interaction terms $\Phi(x_i,x_j)$ as the following equation:
$$\Phi(x_i)=\Phi(x_i,x_i)+\frac{1}{2}\sum_{j\neq i}\Phi(x_i,x_j)$$
An interaction effect terms $\Phi(x_i,x_j)$ is defined as follows:
\begin{multline}
\Phi(x_i,x_j) \\ = \sum_{S\subseteq\{1,\cdots,K\}/\{i,j\}} \frac{2\cdot|S|!(K-|S|-1)!}{K!} [f_x(S\cup\{i,j\}) \\ - f_x(S\cup\{i\})-f_x(S\cup\{j\})+f_x(S)]
\end{multline}
A main effect terms $\Phi(x_i,x_i)$ is defined using the interaction terms $\Phi(x_i,x_j)$ and the Shapley values $\Phi(x_i)$.
$$\Phi(x_i,x_i)=\ \Phi(x_i)-\frac{1}{2}\sum_{j\neq i}\Phi(x_i,x_j)$$
Using the Shapley-Taylor index, the three-variable function $F(x,y,z)$ in Eq.~\ref{3value} is evaluated as $\Phi_{x,x}=f(x)$, $\Phi_{x,y}=g(x,y)$ and $\Phi_{x,z}=g(x,z)$.
The Shapley-Taylor index enables to evaluate the effect of the main effect term and the interaction terms separately.

The Shapley interaction value~\cite{lundberg2020local} is similar to the Shapley-Taylor index; however, the Shapley interaction value cannot separate interactions clearly, i.e $\Phi_{x,x} \neq f(x)$, $\Phi_{x,y} \neq g(x,y)$ and $\Phi_{x,z} \neq g(x,z)$.

\section{Experiments}
We introduce a method for interpreting machine learning models by separating the effects of single features on outcomes from the interactions among features.
The method is applied to the cancer cohort data of Kyushu University Hospital (N=29,080) to analyze what combination of factors contributes to the risk of colon cancer.

\subsection{Study subjects}
We obtained data from the electronic medical record (EMR) system at Kyushu University Hospital (Japan) for 311,391 patients between January 1st, 2008 and December 31th, 2017.
This practical care information included age, sex, height, weight, smoking status, diagnoses [International Classification of Disease version 10 (ICD-10) codes], laboratory test results, and details of prescription medications.
Eligible patients were 20 to 69 years old (n = 203,104) and had recorded serum bilirubin levels (n = 108,014).
In addition, the patients, who had a history of admission and were followed up for over 1 year, were included in the analysis to increase the accuracy of their information (n = 41,415).
Patients were excluded if they had a previous history of cancers or had ICD-10 codes corresponding to liver cirrhosis or hemolytic anemia or had other hepatobiliary diseases with abnormal liver enzyme levels.
Cancer cases diagnosed within 1 year from recruitment into the study were excluded to minimize potential reverse causality.
In addition, patients with serum bilirubin levels over 2.0 mg/dL were excluded because those patients may have had unidentified pathological  conditions affecting serum bilirubin levels, although some of them had hereditary hyperbilirubinemia, such as Gilbert's syndrome.
Finally, a total of 29,080 subjects (12,946 men and 16,134 women) were eligible for inclusion in the analysis~\cite{inoguchi2021association}.
Table~\ref{baseline} shows the baseline characteristics of the study subjects. The median age was 52 years old, and the median follow-up time was 4.7 years.
There were 315 colon (173 men, 142 women) cases in this study.

\begin{table}[t]
\caption{Baseline characteristics of the study subjects. Data was expressed as median (IQR) or mean (SD). The number (No.) was expressed as absolute value and \%.}
\label{baseline}
\vskip 0.15in
\begin{center}
\begin{small}
\begin{sc}
\begin{tabular}{c|ccc}
\toprule
 & Total & Men & Women \\
\midrule
No & 29,080 & 12,946 & 16,134 \\ \hline
Follow-up & 4.7 & 4.6 & 4.8 \\
{}[years] & (2.4-7.9) & (2.3-7.7) & (2.5-8.0) \\ \hline
Age & 52 & 55 & 49 \\
 & (37-62) & (42-63) & (34-60) \\ \hline
BMI[kg/m2] & 23.0 (4.2) & 23.7 (3.8) & 22.4 (4.3) \\ \hline
%kg/m2 & (n=28105) & (n=12532) & (n=15573) \\
BIL & 0.65 & 0.70 & 0.60 \\
{}[mg/dL] & (0.50-0.82) & (0.54-0.90) & (0.50-0.80) \\ \hline
Smoking & 8301 (32\%) & 6021 (52\%) & 2280 (16\%) \\ \hline
%& (n=26144) &  (n=11576) &  (n=11576) \\
Diabetes & 5120 (18\%) & 2958 (23\%) & 2162 (13\%) \\
\bottomrule
\end{tabular}
\end{sc}
\end{small}
\end{center}
\vskip -0.1in
\end{table}

\subsection{Analysis Method}
A survival time analysis using the Cox proportional hazards model was conducted with the seven baseline characteristics as features and the development of colon cancer as the outcome.
In general, logistic regression is often applied as the baseline hazard function in the Cox proportional hazards model; however, the Gradient Boosting Decision Tree (GBDT), a typical machine learning model method, was applied in this analysis.
The baseline hazard function generated by the GBDT is analyzed by SHAP to determine what combination of baseline characteristics contributes to the development of colon cancer.

\section{Results}
\subsection{Results by existing method}
\begin{figure}[ht]
\vskip 0.2in
\begin{center}
\centerline{\includegraphics[width=\columnwidth]{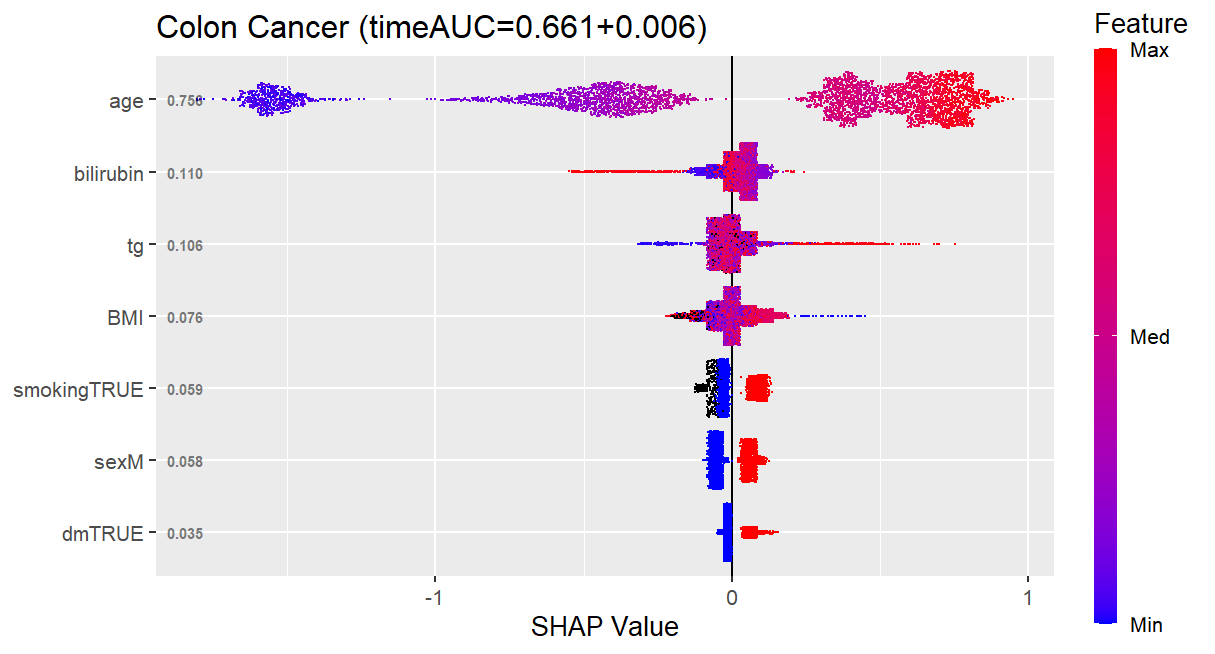}}
\caption{SHAP summary plots for colon cancer risk}
\label{shap_summary}
\end{center}
\vskip -0.2in
\end{figure}

To validate the predictive accuracy of the predictor, we drew a time-dependent ROC of the prediction model and found the 10-year average of its cross-validation AUC was 0.661.
Figure~\ref{shap_summary} shows SHapley Additive exPlanation (SHAP) summary plots for colon cancer.
In the plot, features are sorted by their importance and stacked vertically.
Each row plot is a summary of the SHAP dependence plot of each feature $X_i$.
Each dot represents a patient's SHAP value $\phi(x_i^{(j)})$ plotted horizontally.
Each dot is colored by the value of the feature, from low (blue) to high (red).  Black dots represent missing values.
If red points are plotted at the lower side and blue dots are plotted at the higher side, then the risk becomes higher as the value increases.
Since a SHAP summary plot shows the importance of feature values and an abstract of the SHAP dependence plot, it is useful for overviewing the SHAP analysis.
 
\begin{figure}[ht]
\vskip 0.2in
\begin{center}
\centerline{\includegraphics[width=\columnwidth]{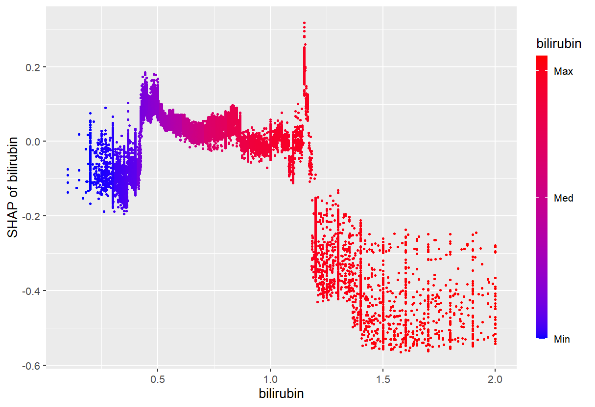}}
\caption{SHAP dependence plots for colon cancer risk against serum bilirubin levels. SHAP values are shown on the y-axis.}
\label{sdp_all}
\end{center}
\vskip -0.2in
\end{figure}

Figure~\ref{sdp_all} is a SHAP dependence plots for colon cancer risk against serum bilirubin levels and shows the SHAP value decreases significantly when bilirubin is larger than 1.2 mg/dl.
This indicates that bilirubin 1.2 mg/dl is the threshold for decreasing the risk of developing colon cancer.
One of the possible causes is as follows.
Cancer-associated infections, smoking, obesity, diabetes, ionizing and ultraviolet radiation, and air pollution are established risk factors for cancer development.
All of these factors are likely to be associated with increased reactive oxygen species (ROS) production in humans.
Increased ROS production has been hypothesized to damage DNA, proteins, and lipids, and thus initiate or promote cancer development.
Since bilirubin is a strong endogenous antioxidant, higher serum bilirubin levels reduce the cancer risk through decreasing of ROS production~\cite{inoguchi2021association}.

In the SHAP Dependence Plot in Figure~\ref{shap_summary}, SHAP values are distributed widely for the same bilirubin value (e.g., SHAP values are distributed widely from -0.4 to -0.2 when bilirubin is 1.2).
This is caused by an interaction between bilirubin and other features.

\subsection{Results by our method}
The Shapley-Taylor Index was used to extract main effects and interactions separately and their importance are evaluated using the standard deviation of the SHAP value~\cite{nohara2022shap}.
Table~\ref{importance_top10} shows the 10 most important features for developing colon cancer.

If we focus only on the main effect of features,  the important features are age, bilirubin, triglycerides, BMI, and smoking history, in descending order.
This result is consistent with the ranking of the existing method in Figure~\ref{shap_summary}.

\begin{table}[t]
\caption{Importance of main and interaction effect terms for colon cancer}
\label{importance_top10}
\vskip 0.15in
\begin{center}
\begin{small}
\begin{sc}
\begin{tabular}{lcccc}
\toprule
rank & Feature1 & Feature2 & importance \\
\midrule
1 & age & age & 0.763 \\
2 & bilirubin & bilirubin & 0.115 \\
3 & TG & TG & 0.097 \\
4 & TG & age & 0.083 \\
5 & BMI & BMI & 0.074 \\
6 & BMI & age & 0.064 \\
7 & smoking & smoking & 0.061 \\
8 & sex & sex & 0.054 \\
9 & bilirubin & age & 0.044 \\
10 & DM & DM & 0.033 \\
\bottomrule
\end{tabular}
\end{sc}
\end{small}
\end{center}
\vskip -0.1in
\end{table}

\begin{figure}[ht]
\vskip 0.2in
\begin{center}
\centerline{\includegraphics[width=\columnwidth]{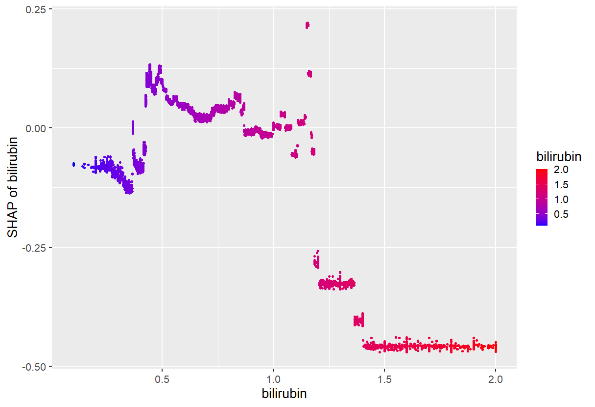}}
\caption{SHAP dependence plots for colon cancer risk evaluating main effect of serum bilirubin levels.}
\label{sdp_single}
\end{center}
\vskip -0.2in
\end{figure}

Figure~\ref{sdp_single} is a SHAP dependence plot for colon cancer risk evaluating the main effect of serum bilirubin levels and shows the SHAP value decreases significantly when bilirubin is larger than 1.2 mg/dl.
This shape is very similar to SHAP Dependence Plot in Figure~\ref{sdp_all}; however, SHAP values are not distributed widely for the same bilirubin value (e.g., all SHAP values are almost -0.3 when bilirubin is 1.2).
This is caused by eliminating the interaction between bilirubin and other features from the SHAP value of bilirubin. 

\begin{figure}[ht]
\vskip 0.2in
\begin{center}
\centerline{\includegraphics[width=\columnwidth]{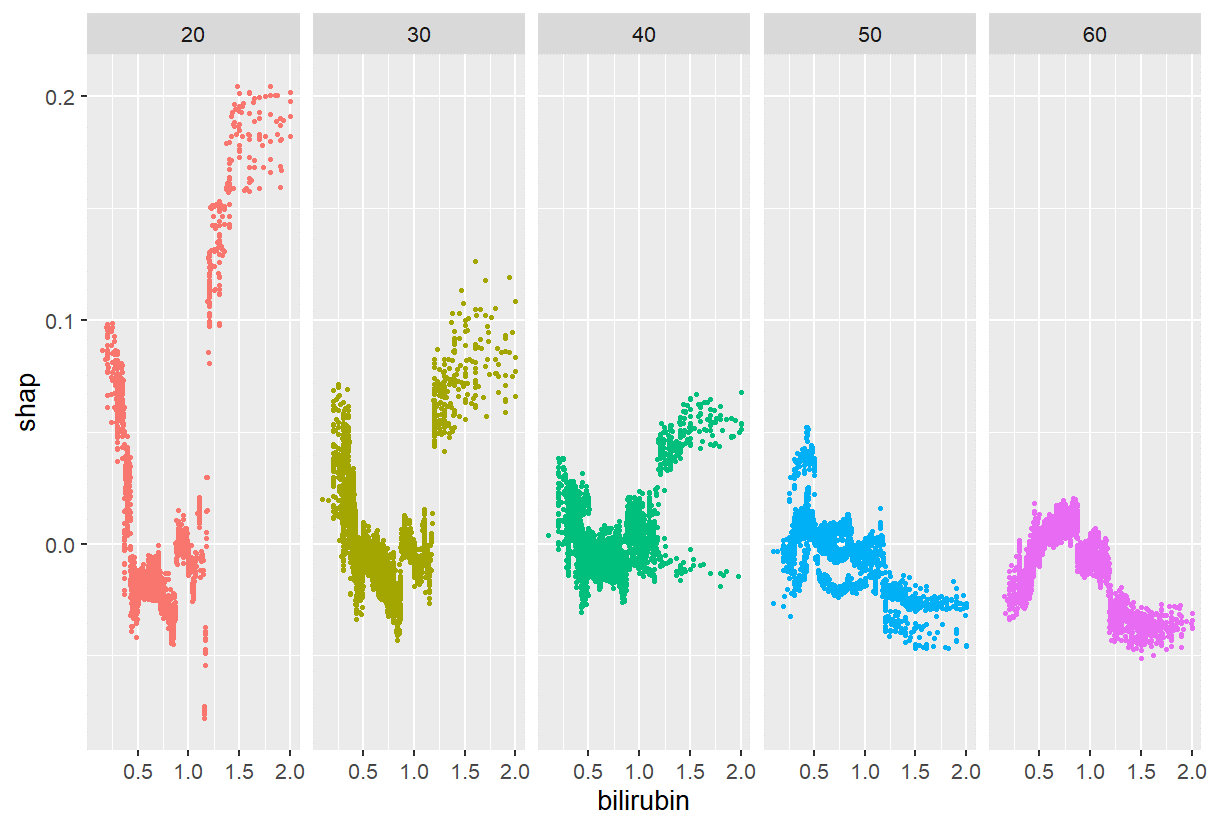}}
\caption{SHAP dependence plots for colon cancer risk evaluating interaction effect of serum bilirubin levels and ages.}
\label{sdp_age}
\end{center}
\vskip -0.2in
\end{figure}

Figure~\ref{sdp_age} shows the SHAP dependence plots evaluating the interaction effect of serum bilirubin levels and ages.
In the group of 20s, we found the interaction effect rises significantly when bilirubin exceeds 1.2 mg/dl, while those in their 50s and 60s show a slight decrease in the interaction value when bilirubin exceeds 1.2 mg/dl.

\begin{figure}[ht]
\vskip 0.2in
\begin{center}
\centerline{\includegraphics[width=\columnwidth]{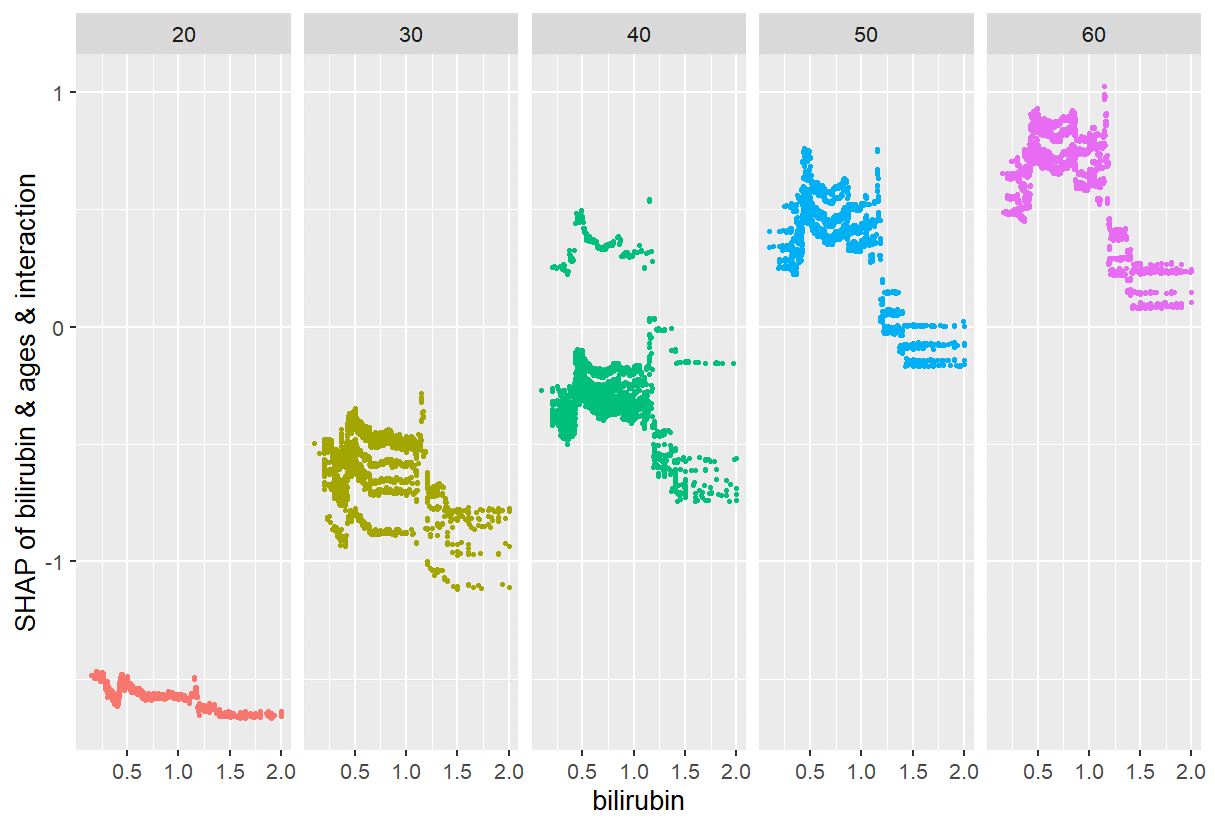}}
\caption{SHAP dependence plots for colon cancer risk evaluating main and interaction effects of serum bilirubin levels and ages}
\label{sdp_age_sum}
\end{center}
\vskip -0.2in
\end{figure}

Figure~\ref{sdp_age_sum} shows the SHAP dependence plots evaluating the main and interaction effects of serum bilirubin levels and ages.
For the 20s, when bilirubin exceeds 1.2 mg/dl, the decrease in the main effect of bilirubin cancels out the increase in the interaction effect, and the sum of their SHAP values does not change significantly.
On the other hand, for the 30s and older, the decrease in the main effect of bilirubin outweighed the increase in the interaction effect, and the sum of their SHAP values decreased when bilirubin exceeds 1.2 mg/dl.
Since the interaction effect is getting smaller with older age, high bilirubin levels decrease the risk of colon cancer, especially for the elderly.
This is because there is much room for risk reduction for the elderly since their risk is larger than that of the young generation. Therefore, high bilirubin levels effectively work for prevention of the cancer, especially for the elderly.

\section{Conclusion}
In this paper, we introduced the Shapley-Taylor index as a method for interpreting machine learning models.
The index separates the SHAP value into the effects of single features and the interactions.
The method is applied to the cancer cohort data of Kyushu University Hospital (N=29,080) to analyze what combination of factors contributes to the risk of colon cancer.
We found high bilirubin levels effectively work for prevention of the colon cancer, especially for the elderly.

\section*{Acknowledgments}
This work was supported by JSPS KAKENHI Grant Number JP20K11938.


\bibliography{SHAPinteraction}
\bibliographystyle{icml2022}

\end{document}